\documentclass[runningheads]{llncs}
\usepackage[T1]{fontenc}
\usepackage{graphicx}
\usepackage[ruled,algo2e]{algorithm2e}
\usepackage{esvect}
\usepackage{array}
\usepackage{amsmath}
\usepackage{booktabs}
\usepackage{amssymb}
\usepackage{hyperref}
\usepackage[most]{tcolorbox}
\begin{document}
\title{Feature Generation Using LLMs: An Evolutionary Algorithm Approach\thanks{This is the author's preprint version of a paper later published as: A. Nourbakhsh, B. Alcaraz, and C. Schommer, ``Feature Generation Using LLMs: An Evolutionary Algorithm Approach,'' in \textit{Advances in Explainability, Agents, and Large Language Models}, CALM 2024, Communications in Computer and Information Science, vol. 2471, Springer, Cham, 2025. DOI: \url{https://doi.org/10.1007/978-3-031-89103-8_4}.}}
%
%
\author{Aria Nourbakhsh\inst{1}\orcidID{0009-0007-2233-3155} \and
Benoît Alcaraz\inst{1}\orcidID{0000-0002-7507-5328} 
\inst{1}\and Christoph Schommer \orcidID{0000-0002-0308-7637} }
\authorrunning{A. Nourbakhsh et al.}
%
\institute{University of Luxembourg, Esch-sur-Alzette, Luxembourg \\
\email{aria.nourbakhsh@uni.lu}}

\maketitle              
\begin{abstract}
A crucial step in machine learning pipelines is to present each entity with features or attributes that are representative of the characteristics of the processed entities. Feature engineering is an important step in finding a relation among attributes that otherwise may not be processed by the ML algorithms. Meanwhile, Large Language Models have shown promising abilities in coding, mathematical reasoning, and processing world knowledge. In this work, we utilize an LLM for the problem of feature generation from tabular data based on the previously given features. We have created a pipeline that takes a set of attributes and a prompt to generate new features. Then, our selection algorithm selects the best-performing sets of attributes. We apply our method to eight datasets from different domains and data types. Our results show that, in most cases, the language model can produce new features based on mathematical and logical operators that are useful for the given tasks and can improve classification results.

\keywords{Feature Generation  \and Large Language Model \and Machine Learning.}
\end{abstract}
\section{Introduction}

In the field of Machine Learning (ML), raw data must be transformed into meaningful features\footnote{Or attributes. Here, we use both terms interchangeably.} to capture relevant patterns effectively. These attributes describe an entity by numerical, boolean, or string values. For optimal categorization into different classes or clusters, features must be distinctive and informative, allowing the algorithm to detect hidden patterns across diverse objects. In other words, ML algorithms induce rules based on how features are represented within the feature space~\cite{motoda2002feature}. The importance of features cannot be overstated. These features capture the underlying structure of data. The most powerful and state-of-the-art algorithms depend on the quality of these features to perform well.



To extract these features from raw data, one needs to create, select, and modify them based on the information that can be retrieved from the entities of the problem. This process is time-consuming and costly as it requires human labor. An approach to mitigate the issue is to generate attributes based on some of the attributes that have already been extracted for representation~\cite{markovitch2002feature}. 

Given a base set of features, another issue is that some of the ML algorithms, such as linear regression, treat each attribute separately. In this process, a hidden relation among the extracted features can be neglected, and hence, we would miss some crucial information. Those relations are often limited by logical and arithmetic operations between the result of comparisons between the value of an attribute and a target value (e.g., `$size = big$', or `$size \geq 3.5$'). For instance, if the pattern to identify corresponds to the sum of two numeric attributes being equal to five, it is hard to create a propositional logic formula being true on this condition when the attribute `sum' is not represented in the data, and this is even less scalable (e.g., for two attributes $a$ and $b$, you may create a rule as $(a = 0 \land b = 5) \lor (a = 1 \land b = 4) \lor ...$, but this formula, already containing six terms in its disjunctive normal form, would become even longer if the sums were depending on three attributes).

Given the number of features for a dataset, a combination of these features with logical and mathematical operators, such as mean or addition, comprises a vast search space, making it not feasible to look for such relations in a brute-force manner. Moreover, the combination of features is unbounded, as one can combine the newly generated features indefinitely. Some of the approaches to tackle the feature generation problem based on the given features are mentioned in the section~\ref{sec:related_word}.

Meanwhile, we see significant advancements in the field of AI due to the emergence of Large Language Models (LLMs). Every day, newer capabilities of these models are explored, and they are applied to different problems. These models are trained on a vast amount of data, and they have the ability to understand, generate, and manipulate human languages with capabilities in solving logical and mathematical problems~\cite{chang2024survey}. More importantly, they have shown their capabilities in generating codes in different programming languages~\cite{coello2024effectiveness,liu2024your}. 

Recently, the scientific community has been exploring the use of LLMs in competitive generation frameworks, where multiple outputs compete to produce superior results~\cite{meyerson2023language,wu2024evolutionary}. This approach, which involves refining outputs iteratively to discover new functions and optimize results, has become central to the \textit{funsearch} method~\cite{romera2024mathematical}, pushing the boundaries of search space exploration.
Building upon the funsearch approach, in this paper, we apply a relatively small language model (LLaMA3.1 7B~\cite{dubey2024llama3herdmodels}) to the problem of feature engineering based on the existing features of a dataset. We experiment with different settings, such as keeping the newly generated features in the prompt, ignoring them, and anonymizing them in a given prompt. Then, we query the language model with the prompt to create functions to be applied to a given dataset to produce new attributes. A selection algorithm selects the better-performing sets of attributes. We apply this method to eight different datasets with two classification algorithms. The results show improvement in most of the settings and datasets.

The advantage of this approach is to a) bypass the combination of all the relations in the possible search spaces and leverage the production of functions by an LLM to compute them. b) Producing transparent features with a function that could be intuitively interpretable. c) Diverse sets of features can be produced by this approach that can be scalable to other tasks and datasets. d) It can take an arbitrary number of features into account, i.e., we do not restrict the feature generation based on a certain number of features. It helps the ML algorithm to solve problems in datasets such as \textit{Monk-2} (see section~\ref{subsec:data}), where the final label is a product of a relation between three attributes. We show that the LLM can take an arbitrary number of features and, without giving any predefined operator, still produce features that improve the predictability of the model. In this process, we use a feature selection algorithm inspired by the genetic algorithm to keep and rank the highest and best set of attributes while still exploring less successful sets of features to avoid being stuck in local optima.

\section{Related Work}
\label{sec:related_word}
In this section, we explore some of the related concepts, such as feature engineering and prompting, and we look at the state-of-the-art approaches to using LLMs in feature engineering for tabular datasets.

\subsection{Feature Engineering}

All the modern data-driven approaches of machine learning, such as Decision Trees (DT), Neural Networks, and linear classifiers (e.g., Support Vector Machines), take a given data point as a vector of features~\cite{6472238}. These features can consist of various common data types used in computer science to represent entities, such as numerical values, boolean flags, or categorical strings. Effective feature extraction and selection techniques enhance the representation of the underlying data structure, improving the performance and generalization of downstream tasks or algorithms~\cite{motoda2002feature,dash1997feature,tang2014feature}. Feature generation and engineering depend on factors like the availability of extractable information and domain knowledge, which can be resource-intensive. However, new features can often be derived from existing ones~\cite{coates2011analysis,heaton2016empirical}. For instance, Body Mass Index (BMI) is calculated as the ratio of weight to height~\cite{heaton2016empirical},  or Total Sales Revenue is the product of $unit\_price   \times quantity\_sold$.

This type of feature construction is important for better classification performance~\cite{dor2012strengthening}. Many approaches to this type of feature construction have been developed through the years, such as greedy strategy~\cite{Shafti10.1145/1068009.1068317},  applying polynomial functions~\cite{sutton1991learning}, and using arithmetic and logical operators to construct new feature~\cite{markovitch2002feature}. For example, in~\cite{dor2012strengthening}, they apply common mathematical functions and arithmetic operators by randomly choosing an operator and applying them to pairs of features. Then, an algorithm keeps the best-performing sets of features over many iterations.

Alternative approaches, such as Reinforcement Learning, have been employed to optimize feature spaces~\cite{khurana2018feature}, and tree traversal techniques have been utilized for feature generation and selection~\cite{khurana2016cognito}. It is also possible to combine a pair or more number of features to create a new one~\cite{katz2016explorekit}.

\subsection{Prompting, Feature Selection, and Generation}

Prompting using LLMs is a new trend that exploits knowledge by predicting the probability of word sequences. These language models show incredible performance in reasoning~\cite{wang2024causalbench,ahn2024large} and domain-specific applications such as medicine~\cite{karabacak2023embracing} and finance~\cite{li2023large}.
However, these models have some shortcomings. Since they are trained on data from the internet and rely on language modeling to predict the most probable tokens, they are prone to hallucinations and generating biased outputs~\cite{navigli2023biases}. 

We identified four relevant studies that utilize LLMs for feature selection and feature engineering.~\cite{jeong2024llm} applies LLMs to feature selection, using various prompts and strategies for selecting, ranking, and incrementally choosing features. Similarly, \cite{li2024exploring} employs prompt-based techniques for feature selection and shows that the feature importance identified by LLMs correlates strongly with traditional feature importance estimation methods. Their approach incorporates feature values alongside target labels. In another setting, they provide the data and feature description to the LLM for feature selection.

For feature engineering~\cite{pmlr-v235-han24f} applies a prompt to generate features by a limited set of mathematical operators. They show promising results by providing the task description, features, and examples of training data. They explicitly ask the LLM about analyzing the relevance of the features and the task at hand. Further, they prompt by asking to create ten binary features by the use of `is in', `$\geq$', and `$\leq$' keywords. Also, they explicitly ask about the possible range of values from the LLM. Notably, they use a limited set of operators and more complex, longer prompts.

The work in `Dynamic and Adaptive Feature Generation with LLM'~\cite{zhang2024dynamic} is the most closely related approach to our work. In their method, a feature set and a predefined set of mathematical operators are provided to a pipeline by prompting an LLM. Then, in prompting, they combine original and newly generated features. This is followed by an iterative feature selection process, which evaluates the generated features on downstream tasks and retains the optimal dataset. The process employs a set of predefined unary and binary mathematical operators.

Our approach differs from the work above in several ways: a) Unlike~\cite{zhang2024dynamic}, we do not restrict the LLM to a predefined set of operators or binary feature combinations. Instead, our method allows the creation of new features using any number of related features, with outputs that can be boolean, string, or real numbers. This contrasts with~\cite{pmlr-v235-han24f}, where output functions are limited to binary values. b) We demonstrate that the LLM can generate new features with a relatively short prompt. c) We show that even with minimal information about the dataset, the LLM can produce new features that enhance classification performance. d)  Our results indicate that competitive performance can be achieved using a local LLM that runs on a personal computer, with only a short prompt describing the intended output and minimal information about the data. e) To mitigate the influence of biased knowledge within LLMs, we propose anonymizing the input and relying solely on the LLM’s ability to generate well-formed, viable functions without any domain-specific knowledge.

\section{Methodology}
\label{Methodolgy}
This section describes the pipeline and the prompting command we devised for our approach.

\subsection{Pipeline}

Our pipeline\footnote{Code is available on the GitHub page: https://github.com/zaap38/funsearch-att} is composed of three modules: 1) A feature generator module powered by an LLM, 2) An evaluator that applies a machine learning algorithm to learn from and assess the generated features, and 3) A selection algorithm that identifies the top-performing feature sets to refine and further apply the generated features. The pseudocode for this pipeline is outlined in Algorithm \ref{alg-1}.

The pipeline gets the raw dataset, The number of iterations $n$, and a maximum sample count $M$. Each sample is a set of attributes. The output of the pipeline is a set of refined dataset samples along with their F1-measures.
First, the algorithm calculates the F1 of the initial dataset $D$, and then a set of samples is initialized, containing the original dataset $D$ and its associated F1-measure. The algorithm then enters a loop that will run for 
$n$ iterations. It first sums the F1s of all current samples. It then selects one sample randomly, with a higher probability of selection for samples with higher F1. This is done by generating a random number between 0 and the sum of all the F1s and progressively subtracting each sample's F1 from this random number until it becomes less than or equal to zero, indicating the selected sample.

Once a sample is selected, the algorithm generates a mutation prompt by calling \texttt{getPrompt} on the selected dataset. This prompt is then passed to an LLM via askLLM method to generate a new attribute. The new attribute is added to the selected sample, and the F1 of this mutated dataset is recalculated. The mutated dataset and its new F1 are added to the list of samples.
The LLM generates both an attribute name and a Python expression, formatted in a way that can be directly applied to a Pandas DataFrame structure (see section \ref{subsec:Prompting}) We iteratively prompt the LLM to create new features using the above mentioned evolutionary process. 

After mutating and adding the new sample, the algorithm cleans up the sample set. It creates a new set of samples by randomly selecting up to 
$M$ samples from the current set, again using the probability-based selection method (where F1-measure samples have a higher chance of being selected).
After selecting 
$M$ samples, the old sample set is replaced with the new, cleaned-up set. The loop repeats this process for $n$ iterations, refining the dataset samples and their associated accuracies. After the loop finishes, the algorithm outputs the final set of samples, each with its associated F1, and we select the highest-scoring sample as the selected dataset.


\begin{algorithm2e}[H]
\footnotesize
\DontPrintSemicolon
\caption{Pseudo-code for dataset mutation and selection}\label{alg-1}
\SetAlgoLined

\KwIn{Dataset $D$, Number of iterations $n$, Maximum sample count $M$}
\KwOut{A set of refined dataset samples with their F1s}

$F1 \gets \texttt{getF1}(D)$\;
$samples \gets \{(D, F1)\}$\;

\For{$i \gets 1$ \textbf{to} $n$}{
    $sum \gets 0$\;
    \ForEach{$sample \in samples$}{
        $sum \gets sum + sample[1]$\;
    }
    $rand \gets \texttt{random}(0, sum)$\;
    \ForEach{$sample \in samples$}{
        $rand \gets rand - sample[1]$\;
        \If{$rand \leq 0$}{
            $d \gets sample[0]$\;
            \textbf{break}\;
        }
    }

    $msg \gets \texttt{getPrompt}(d)$\;
    $attribute \gets \texttt{askLLM}(msg)$\;
    $d \gets \texttt{addAttribute}(d, attribute)$\;
    
    $F1 \gets \texttt{getF1}(d)$\;
    $samples.\texttt{add}((d, F1))$\;

    $new\_samples \gets \emptyset$\;
    \For{$j \gets 1$ \textbf{to} $M$}{
        $sum \gets 0$\;
        \ForEach{$sample \in samples$}{
            $sum \gets sum + sample[1]$\;
        }
        $rand \gets \texttt{random}(0, sum)$\;
        \ForEach{$sample \in samples$}{
            $rand \gets rand - sample[1]$\;
            \If{$rand \leq 0$}{
                $new\_samples.\texttt{add}(sample)$\;
                \textbf{break}\;
            }
        }
    }
    $samples \gets new\_samples$\;
}
\end{algorithm2e}

\subsection{Prompting}\label{subsec:Prompting}
At the heart of our pipeline, there is the LLM that gets a prompt and produces a Pythonic expression applicable to a Pandas Dataframe~\cite{reback2020pandas}.
We use the following prompt shown in Figure \ref{fig:prompt} to instruct the LLM on the task it needs to perform. Note that the LLM is not aware of the previous prompts (i.e., we don't have a memory of the previous prompts for the LLM).


        


\begin{figure}[htbp]
\centering
\begin{tcolorbox}[colframe=blue, colback=blue!10, coltitle=black, sharp corners=southwest, rounded corners, width=0.9\textwidth]
\ttfamily
Take the following dataframe attributes, their dtype, and possible values. Create exactly one new attribute based on them using mathematical and logical operators. You can use numpy and math libraries if needed. return the result without code format and without extra text as
below:
\\
\\
n: new\_attribute\_name\\
a short expression that can be applied to a dataframe
\\
\\
\\
df[feature\_name\_1] : dtype-possible values [...]\\
.\\
.\\
.\\
df[feature\_name\_n] : dtype-possible values [...]
\end{tcolorbox}
\caption{The prompt used to generate a Python code to create a function applicable to a Pandas dataframe.}
\label{fig:prompt}
\end{figure}

In the prompt, we provide the original feature names and their data type. For categorical and boolean values, as long as they are fewer than 20 values, we enumerate their possible values. For the attributes that contain integers or floating point values, we give the minimum and maximum range of that column to the prompt. 

In our experiments, we apply different settings: a) We retain the newly generated feature sets and provide the LLM with both the original and newly generated features. The advantage of this approach is the potential to discover valuable features derived from the newly created ones. However, for large datasets, this can increase the model’s computational complexity and running time. Also, having more features can lead to the `curse of dimensionality'~\cite{bellman1957dynamic}.

b) The model ignores the newly generated features, creating only one new attribute based on the original dataset. This approach reduces evaluation time and minimizes the risk of introducing noise into the feature sets. It also reduces the chance of the LLM creating malformed functions.

c) Is the same as (b); however, we anonymize the feature names and values so the LLM cannot use any learned knowledge or correlations to build new features.

More formally, our approach can be expressed as follows: Let $A$ be all the possible attributes. We define the function $P : 2^A \rightarrow \Sigma^\mathbb{N}$ (where $\Sigma$ is the alphabet containing all the ASCII characters, and so $\Sigma^\mathbb{N}$ is the formal definition of a string), which from the list of the attributes of a dataset, returns a string which we will use as our prompt for the LLM. Consequently, let $V$ be all the dataset rows and $V_i$ be one row. We define the LLM as a function $L : \Sigma^\mathbb{N} \rightarrow \mathcal{M}$ where $\mathcal{M}: V_i \rightarrow V_i \cup Y$ with $Y \in \{\mathbb{R}, \Sigma^\mathbb{N}\}$ (meaning it can be a boolean, a real, or a string) and $V_i \in V$.

In scenario (a), $A$ is defined as the set of both the original attributes in the dataset and the newly generated ones. In contrast, in scenario (b), $A$ refers solely to the base set of original attributes.

\section{Evaluation}

This section presents an overview of the datasets and experimental setup, followed by an analysis of the results.

\subsection{Data}
\label{subsec:data}

As mentioned in the introduction, the search space for new attributes can be unbounded. While this may initially seem arbitrary, we envision that an LLM could help narrow the search space by being provided with minimal information about a dataset to produce a mathematical or logical function. For this reason, we chose two types of datasets from previous work and publicly available datasets. The first group of these datasets has rich feature names, so the LLM has access to the context and semantics of the datasets. The second group has three datasets whose feature names do not provide any information about the task.

In a separate experimental setup, for comparison, we also anonymize the first group of the datasets, where each attribute is transformed into `attribute\_n', with `n' denoting the attribute's position in the dataset. Additionally, categorical values are converted into numerical representations. We expect that when the LLM is provided with semantic context of the features, it may be able to generate more insightful new attributes, albeit with potential biases.

We take three out of four datasets used by~\cite{zhang2024dynamic} to compare our approaches. One of their datasets is the Amazon Commerce Review~\cite{amazon_commerce_reviews_215}, which we think is inappropriate for this task as it fits other Natural Language Processing techniques better because each column represents a letter with more than 10000 features. One of the other three datasets is Ionosphere (Ion)~\cite{misc_ionosphere_52}. It has 34 features that represent a phased array of 16 high-frequency antennas. As a result, all the features have numerical values, and the labels are binary for \textit{good} and \textit{bad} signals.

Another dataset is the Diabetes Health Indicators Dataset (Dia)~\cite{teboul2022diabetes}. We take a subset of this dataset. The dataset has 22 health-related features with two labels indicating whether or not a person has diabetes. The dataset has a mix of boolean and real numbers for feature values. 

Abalone is another dataset from the previous work that we used for the experiments~\cite{misc_abalone_1}. Abalone (Aba) has eight features for predicting the age of abalones. These features consist of length, diameter, and other indicators of abalones' characteristics \footnote{This dataset is made for predicting the age of abalones, which has 30 different values. However, in~\cite{zhang2024dynamic}, they indicate that it is a binary prediction, which is different from the original description of data.}.

In addition to the datasets mentioned above, we selected additional datasets spanning various tasks and domains to ensure a broader and more diverse evaluation. The first one, Monk-2~\cite{dataset_monk}, is a synthetic dataset consisting of six numerical attributes with integer values ranging from $1$ to $4$, and two classes ($0$ and $1$). In the specific case of Monk-2, the class has the value $1$ if and only if exactly two of the six attributes have the value $1$. More formally:

$$\text{class} = \begin{cases}
    1 & \text{if } |\{a \in \text{Attributes} \mid a_{\text{value}} = 1\}| = 2 \\
    0 & \text{otherwise}
\end{cases}$$

This dataset is interesting as the class is not bounded to a simple equality/inequality between some attributes and a constant. There exists a trivial propositional formula representing this condition, but it is large enough so it cannot be easily represented with a predefined list of operators containing only the boolean and arithmetic operators or basic functions such as $x^2$, $log_2(x)$, or $\sqrt{x}$. 

Monk-3~\cite{dataset_monk} is similar to Monk-2, but it follows a more complex relationship, and the labels are determined by the following expression: $(a5 = 3 \land a4 = 1) \lor (a5 \neq 4 \land a2 \neq 3)$ where $an$ is a column in the dataset.

Another dataset is named Car dataset~\cite{dataset_car}. It classifies a car in terms of buying acceptability (Unacceptable, Acceptable, Good, or Very good) based on several attributes, such as the number of seats or the size of the luggage boot. This dataset is interesting because it is simple to run a qualitative evaluation of the generated attributes (As it is easy to understand why a low buying price or a large number of seats pushes the acceptability level toward Very good).

Finally, we use Wine~\cite{wine_109} and Predict Students' Dropout and Academic Success (Stu) datasets ~\cite{misc_predict_students'_dropout_and_academic_success_697} with 13 and 36 features respectively.  
Stu dataset contains a diverse set of 
personal, educational, and demographic attributes. The Wine dataset consists of attributes that describe a wine characteristic. These datasets represent distinct domains and feature a variety of attributes and data types, making them well-suited for evaluating our pipeline and the capabilities of the LLM.

\begin{table}[htbp]
\centering
\caption{Datasets size and number of features.}
\label{table:Datasets}
\resizebox{0.9\textwidth}{!}{%
\begin{tabular}{|l|c|c|c|c|c!{\vrule width 1.5pt}c|c|c|}
\hline
\textbf{Dataset}      & \begin{tabular}[c]{@{}c@{}}Abalon \\(Aba)\end{tabular} & Car  & \begin{tabular}[c]{@{}c@{}}Diabetes\\  (Dia)\end{tabular} & \begin{tabular}[c]{@{}c@{}}Student \\ (Stu)\end{tabular} & Wine & Monk-3 & Monk-2 & \begin{tabular}[c]{@{}c@{}}Ionosphere\\  (Ion)\end{tabular} \\ \hline
\textbf{No. Features} & 8            & 6    & 21                                                        & 36                                                       & 13                                                            & 6     & 6   & 34    \\ \hline
\textbf{Size}         & 4178         & 1728 & 14000                                                     & 4425                                                     & 1600                                                          & 170   & 123 & 352  \\ \hline
\end{tabular}
}
\end{table}

Table \ref{table:Datasets} summarizes the number of samples and features in each dataset. The datasets on the right side of the bold line are without meaningful feature names.

\subsection{Results}

This section presents the results of our classification pipeline using the Multi-Layer Perceptron (MLP) and Decision Tree (DT) algorithms. We report four outcomes for each algorithm based on the methodology described in section~\ref{Methodolgy}. First, as a baseline, we provide the results of running the algorithms on the raw, original attributes, referred to as \textit{Raw} method. Next, the results for aggregating both base and generated attributes for prompting are reported under the name \textit{Agg}. In the third scenario, we use only the base attributes for feature generation, also named the \textit{Skip} method.  Lastly, the anonymization of attributes and their values is presented under the label \textit{Anon}.

In Table~\ref{table:result}, we separate the three datasets at the bottom of the table that, by design, are anonymous, with feature names that do not carry any semantic meaning related to the task. For our experiments, we tested various local open-source language models and selected LLaMA3.1 7B~\cite{dubey2024llama3herdmodels} due to its ability to consistently provide better, more diverse, and more parsable functions. All experiments were conducted on a single NVIDIA GeForce 4090 using the scikit-learn~\cite{scikit-learn} implementation of the aforementioned algorithms. Table~\ref{table:result} presents the outcomes of $n=400$ iterations of the pipeline, with an early stopping set to the patience of 20 queries (i.e., if 20 consecutive queries to the LLM do not improve the result on the validation set, the process stops, and the best-performing sample is returned). We split the data into 70\% for training, with the remaining 30\% equally divided between development and testing sets. Following standard supervised learning practices, we optimize the pipeline on the development set and report the results on the test set. Due to the stochastic nature of the LLMs and our pipeline, we report the average result on the test sets and the final best-performing attributes over three runs.


\begin{table}[htbp]
\centering
\caption{The F1 result of our evolutionary pipeline using LLM feature generation and selection algorithm. \textit{Raw} is the baseline without generated features. In \textit{Agg}, we use generated features in the prompt, while in the \textit{Skip}, only the base features are used. \textit{Anon} is the same as \textit{Skip}, but the features are anonymized. The three bottom datasets are anonymous by design. We report the change in results with the best-performing setting compared to the baseline.}
\label{table:result}
\begin{tabular}{c|ccccc|ccccc}
\hline
           & \multicolumn{5}{c|}{DT}                   & \multicolumn{5}{c}{MLP}                   \\ \cline{2-11} 
           & Raw & Agg   & Skip  & Anon & change & Raw & Agg   & Skip  & Anon & change \\ \hline
Aba     & 17.39    & 18.69 & 20.09 & \textbf{20.99} & 3.60    & 24.50     & \textbf{25.08} & 23.07 & 23.98 & 0.58   \\
Car        & 98.07    & \textbf{98.59} & 97.42 & 97.02 & 0.52   & 88.91    & 96.37 & \textbf{98.04} & 88.95 & 9.13   \\
Dia       & 64.38    & \textbf{66.31} & 65.11 & 65.38 & 1.93   & \textbf{71.65}    & 71.00    & 71.02 & 70.38 & -0.63   \\
Stu    & 68.81    & 68.16 & 67.70  & \textbf{69.28} & 0.47   & 72.94    & 73.55 & \textbf{74.58} & 73.20  & 1.64   \\
Wine       & 62.97    & \textbf{63.36} & 61.02 & 61.16 & 0.39   & 60.60     & \textbf{64.81} & 60.35 & 62.66 & 4.21   \\ \hline
Ion     & 88.94    & \textbf{92.61}  & \textbf{92.61} & -    & 3.67   & \textbf{92.32}    & 91.64 & 92.30  & -    & -0.02  \\
Monk-2      & 61.54    & \textbf{80.83} & 72.08 & -    & 19.29  & 62.77    & 68.89 & \textbf{69.04} & -    & 6.27   \\
Monk-3      & 89.46    & \textbf{98.25} & 93.02 & -    & 8.79   & \textbf{94.77}    & \textbf{94.77} & 89.47 & -    & 0 \\
\bottomrule
\end{tabular}
\end{table}

For the DT algorithm, the Aba dataset gets the highest change for the datasets at the top of the table. This is achieved in \textit{Anonym}
setting where the description of the dataset is anonymized. Moreover, we see a great improvement for the anonymous datasets at the bottom of the table, suggesting a great potential for LLMs to produce functions without access to the semantics of the dataset. In particular, for Monk datasets, we see a big jump from the baseline. This shows that the produced attributes could take a ternary relation into account. In other cases, where the model had access to the feature names, keeping the generated features in the feature generation queries has resulted in better scores.

For the MLP algorithm, feature aggregation (\textit{Agg}) and feature skipping (\textit{Skip}), in most cases, improve the performance. This is especially apparent in datasets like Car, Wine and Monk-2, where the \textit{Agg} and \textit{Skip} configurations yield significant performance boosts. Notably, the Stu and Monk-2 datasets still exhibit strong performance even in the anonymized setting, indicating that the algorithms can maintain robustness despite removing semantic information from the feature names.

We were expecting that keeping the generated features to produce new samples (i.e., prompting with the aggregated features) would degrade the quality of the intake and generation of the new features. As we can see for MLP, this behavior is random and based on the produced features; in some cases, the \textit{Skip} method scores better than \textit{Agg} and vice versa. Overall, in 13 out of 16 experiments with DT and MLP, our evolutionary feature generation increases the performances of the results.

Unfortunately, we could not reproduce the results of~\cite{zhang2024dynamic}, even for the base cases without additional features. This discrepancy may stem from differences in random sampling or the splitting of data into training, development, and test sets. Consequently, a direct comparison between our results and theirs is not feasible. Nonetheless, for the Dia dataset, our approach demonstrates a more significant improvement between the raw data and our best-performing model. Specifically, using the DT algorithm, we achieved a 1.93\% increase in F1 measure (from 64.38\% to 66.31\%), in contrast to the marginal decline reported by Zhang et al. (from 59.7\%  to 60\%).

\section{Discussion}
In this section, we look at some insights and behavior of the LLM and the implemented algorithm for optimization.

\subsection{Operators}
\begin{figure}[htbp]
    \centering
    \includegraphics[width=\linewidth]{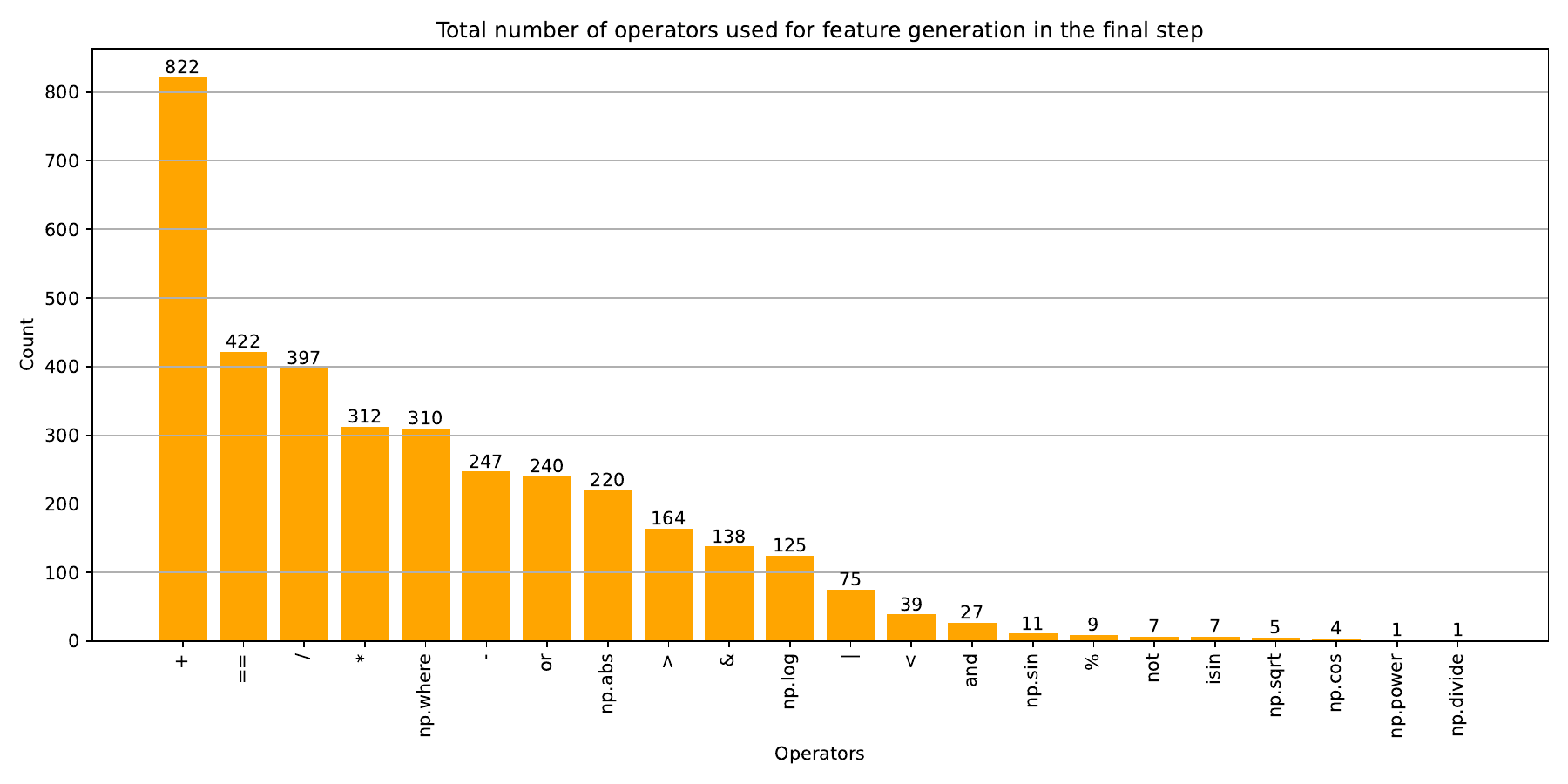} 
    \caption{Figure shows the total number of generated operators in all the final created sets of attributes for all the datasets and runs.}
    \label{fig:operator_count}
\end{figure}

The analysis of the number of operators produced in all runs and datasets, as shown in Figure \ref{fig:operator_count}, reveals that the language model tends to produce arithmetic operators more frequently. Addition ($+$), division ($/$), and multiplication (*) are the most commonly generated operators for the functions. Surprisingly, the model has generated more `$\&$' symbols than the keyword `$and$'. Additionally, there is a preference for using the numpy package over Python's math library.

Logical and comparison operators, such as equality (==), greater than (>), and logical conjunctions (and, or), exhibit significant frequencies, emphasizing their importance in conditional statements and control flows within the generated code. These findings underscore the model's capability to generate code that is not only mathematically robust but also aligned with common programming practices in data science and numerical analysis. However, to force the model to produce arithmetic and mathematical operators, in the prompt (Figure \ref{fig:prompt}), we explicitly  specified, `You can use numpy and math
library if needed.'  Otherwise, the model tends to use operators with boolean outcomes.

\subsection{Number of operators per function}

\begin{figure}[htbp]
    \centering
    \includegraphics[width=\linewidth]{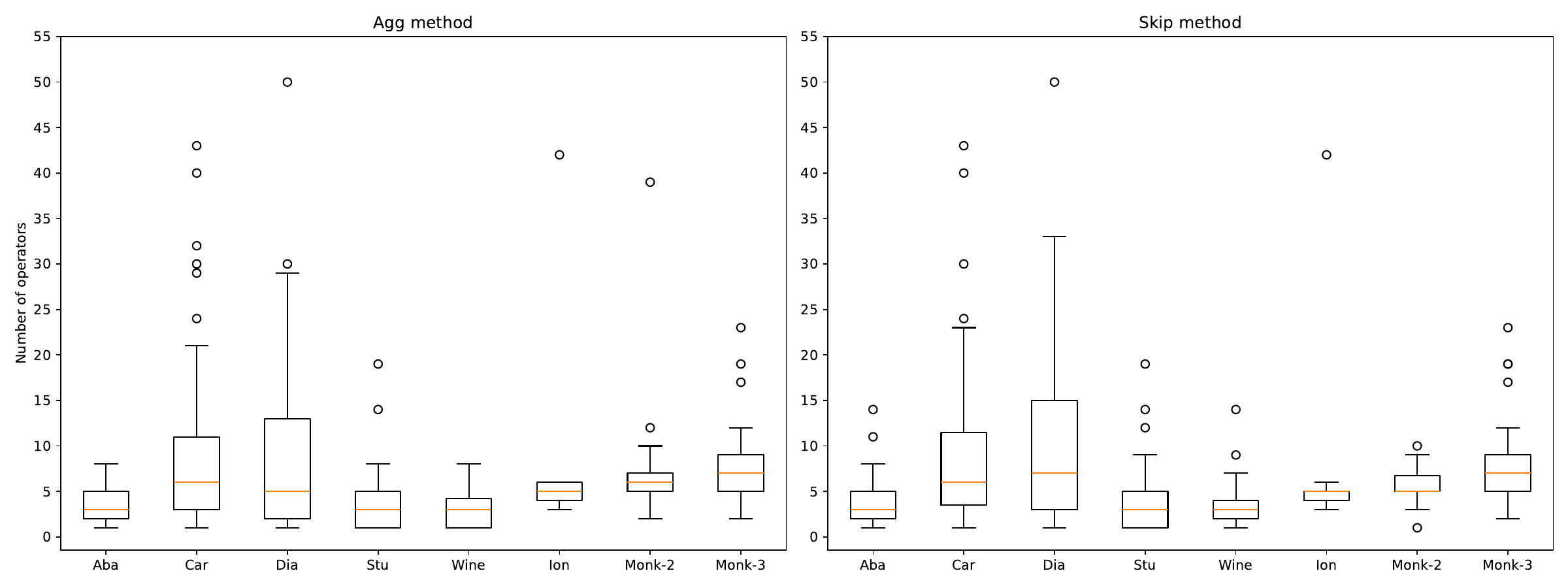} 
    \caption{The boxplot of the distribution of operators per generated functions. In some cases, we see functions with more than 20 operators.}
    \label{fig:operator_per_feat}
\end{figure}

Another interesting measure would be to see how many operators the LLM has generated to generate a new feature. We take the \textit{Agg} and \textit{Skip} methods and visualize the distribution in Figure \ref{fig:operator_per_feat}. The LLM utilizes a broad range of operators, varying from 2 to 50 per function. However, the average number of operators typically falls within the range of 4 to 7. Moreover, we see more and broader range of operators were generated for the Car and Dia datasets. It is an unexpected behavior as the number of base features of these two datasets is noticeably different.


We also conducted Mann-Whitney U tests to compare the number of operators per generated function between the \textit{Agg} and \textit{Skip} methods. In the unadjusted analyses, Dia and Monk-2 showed nominally significant differences ($p = 0.0071$ and $p = 0.0021$, respectively). The early-stopping mechanism in our pipeline limited the accumulation of generated attributes. In experiments without this constraint, as the number of available attributes increased, the LLM tended to generate longer and more complex expressions. These expressions were more likely to fail during Python parsing.

\subsection{Running time}
\begin{figure}[htbp]
    \centering
    \includegraphics[width=\linewidth]{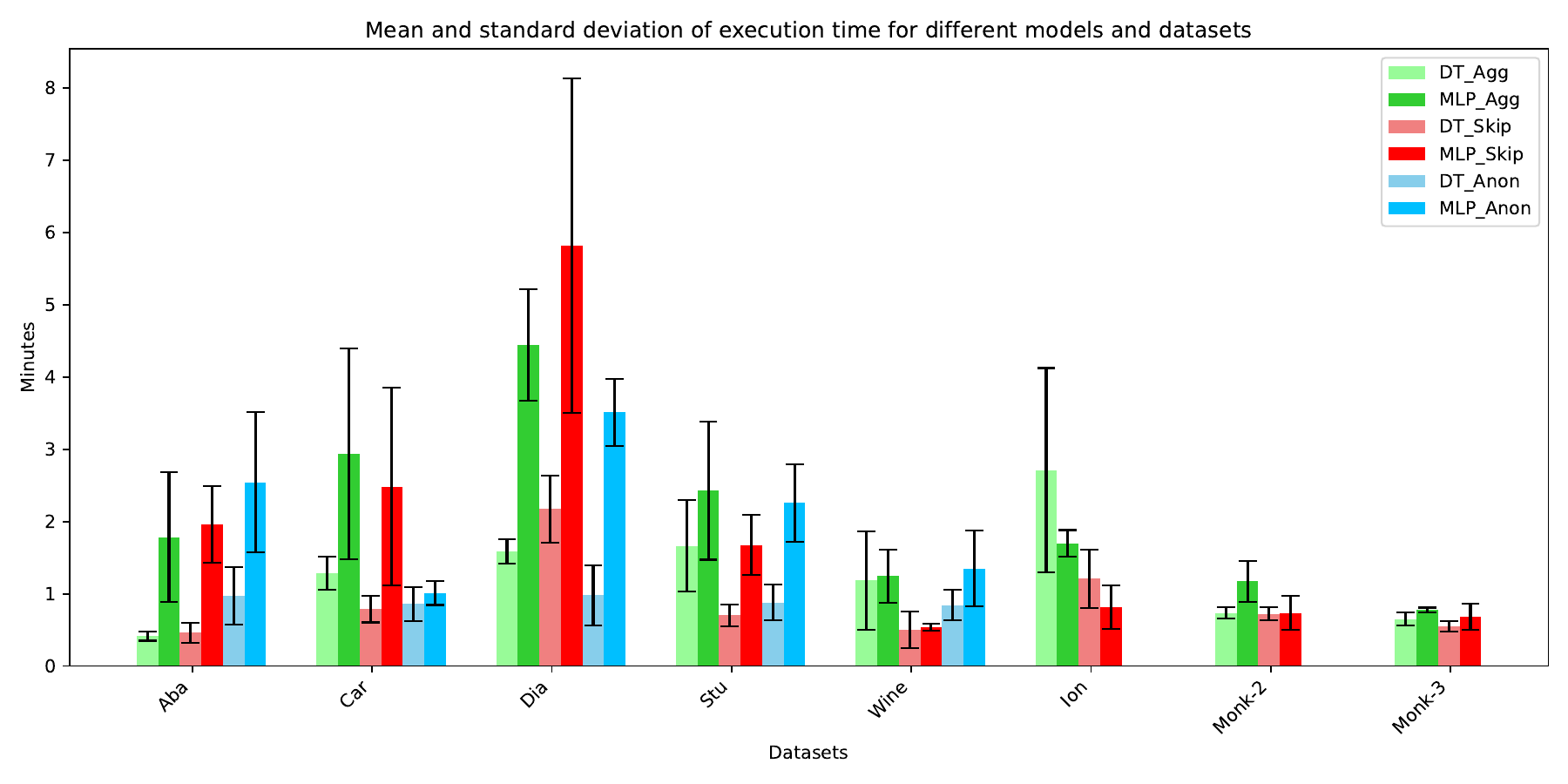} 
    \caption{Figure shows the average and standard deviation of the running time of each approach on each dataset. Please pay attention that the last three datasets are anonymous datasets by design.}
    \label{fig:operator_time}
\end{figure}

Figure \ref{fig:operator_time} shows the average running time of each method on the datasets.
Running time is subjective and depends on hardware, operating system, and other uncontrollable factors. However, on a computer with AMD Ryzen 7900X and Nvidia RTX4090 GPU, we report the average running time of the pipeline over three runs. Also, a model may have converged earlier than the others. According to Figure \ref{fig:operator_time}, the trend shows that, as expected, the larger dataset (Dia) took more time to be processed with our pipeline, while smaller datasets like Monk-2 and Monk-3 were processed more quickly. Also, generally speaking, running the pipeline with the MLP algorithm takes more time than the DT, adding a computational bottleneck to evaluate the features on the downstream tasks. On three datasets, namely Aba and Dia, keeping the generated features(\textit{Agg} models, green bars) has led to a faster convergence\footnote{in our experiments, we tried running the pipeline without the early stopping. In the case of \textit{Agg}, feeding the LLM with newly generated features leads to more error-prone generation of functions.}.

\section{Conclusion}
In this paper, we took the concept of funsearch~\cite{romera2024mathematical} and applied it to the feature generation problem. Namely, we took eight datasets and asked an LLM to generate novel features based on the given attributes. We created three settings: a) We kept the base attributes and added the newly generated ones to the prompt. b) We skipped the newly generated attributes in prompting the LLM, and c) We did the same as (b) with anonymizing the feature names and possible value ranges. On top of that,  We devised an evolutionary algorithm inspired by genetic algorithm, which selects the best-generated sets of attributes.

In this process, we showed that we could reach a fast convergence by utilizing a relatively small local language model that, in most cases, improves the baseline. More importantly, we showed that in the cases where the LLM has no access to contextual knowledge by anonymizing the description of the data, it is still capable of producing functions that may improve the result. This is important because we do not rely on biased knowledge of LLM, and the model only searches for the possible space of the functions to produce new attributes. In other words, our pipeline relied only on the code-generation ability of the LLM.

\section{Limitations and future work}

Our approach has some limitations worth noting. First, we use a small LM. Better results and more parsable functions are expected to be obtained from using a larger LM or a commercial one. 

The whole pipeline and the stochastic nature of the LLMs, make it hard to reproduce the result of each run of the feature generation. Moreover, LLMs are trained on human data, and they are biased toward the training data artifacts. Our suggestion to bypass this problem was to anonymize the description of data in the given prompt. However, it is worth noting that the code-generation ability of the LLM may also be biased, and the LLM tends to generate one operator more commonly than the others. One potential solution to this problem is to explicitly ask the LLM which operators it can use or to specify the returned data types of the function.

\begin{credits}
\subsubsection{\ackname} 
We thank the Luxembourg National Research Fund (FNR) for the funding of this research as part of the project C21 - Collaboration 21:\\
IPBG2020/IS/14839977/C21.

\subsubsection{\discintname}
    None of the author of the paper is having a conflict of interest.
\end{credits}
%
%
%

\bibliographystyle{splncs04}
\bibliography{references}
\end{document}